\documentclass[sigconf]{acmart}

\AtBeginDocument{%
  }
\usepackage{booktabs}
\usepackage{tabularx}
\usepackage{xcolor}
\usepackage{colortbl}
\usepackage{makecell}

\definecolor{ctx}{RGB}{220,235,250}
\definecolor{adp}{RGB}{250,225,220}
\definecolor{rowA}{HTML}{F5F5F5}     
\definecolor{rowB}{HTML}{FFFFFF}     
\definecolor{headbg}{HTML}{E5E5E5}   
\definecolor{taskaccent}{HTML}{2B2B2B}  

\newcommand{\Uturn}{\textcolor{black!75}{\textbf{U:}}}
\newcommand{\Aturn}{\textcolor{black!55}{\textbf{A:}}}

\copyrightyear{2026}
\acmYear{2026}
\setcopyright{cc}
\setcctype{by-nc-nd}
\acmConference[RecSys '26]{20th ACM Conference on Recommender Systems}{September 27-October 02, 2026}{Minneapolis, MN, USA}
\acmBooktitle{20th ACM Conference on Recommender Systems (RecSys '26), September 27-October 02, 2026, Minneapolis, MN, USA}
\acmDOI{10.1145/3773078.3831910}
\acmISBN{979-8-4007-2284-4/2026/09}

\begin{document}

\title{Bootstrapping Conversational Recommendation Agents At Spotify: Synthetic Data Generation and Self-Improvement Loops}

\author{Enrico Palumbo$^1$, Alexandre Tamborrino$^2$, Victor Ode$^3$, Ben Lacker$^3$, Adrià Casas Escoda$^3$, \\Jeremy Hopple$^3$, Marcus Better$^3$, James Leoni$^4$, Hugo Galvão$^3$, Hugues Bouchard$^5$, \\Mounia Lalmas$^4$, José Luis Redondo García$^5$, Abenezer Abebe$^3$, Ann Clifton$^3$, \\Anton Blomberg$^6$, Henrik Lindström$^6$, Dani Doro$^6$, Christine Doig Cardet$^5$}
\affiliation{
{Spotify}\\
\country{$^1$Italy, $^2$France, $^3$USA, $^4$UK, $^5$Spain, $^6$Sweden}
}
\email{enricop@spotify.com}  

\renewcommand{\shortauthors}{Palumbo et al.}
\renewcommand{\shorttitle}{Bootstrapping Conversational Recommendation Agents At Spotify}
\acmArticleType{Research}


\begin{abstract}

Conversational recommendation agents are emerging as a new paradigm for content discovery, enabling users to express complex intents through natural language (e.g., \emph{``recommend Italian indie artists I haven't heard before'' or  ``explain why they fit my taste''}). 
A central challenge in building such agents is optimizing agent planning, i.e., deciding how to select, sequence, and invoke tools. This challenge is particularly acute in cold-start settings, where real user interactions are not yet available.

We introduce a pipeline for multi-turn synthetic data generation and a self-improvement loop to address the lack of interaction data and the difficulty of optimizing agent planning in cold-start settings. The synthetic data pipeline transforms single-turn prompts into realistic multi-turn user–agent conversations, enabling systematic evaluation of conversational capabilities before launch. The self-improvement loop then uses this data and evaluation feedback to combine variance-based contrastive optimization with iterative refinement through a coding agent, allowing the system to automatically identify and fix planning and tool-use errors.

Our approach provides fine-grained insights into conversational capabilities, uncovers issues before deployment, and improves quality by +8\% on top of a highly optimized manual prompt, automatically resolving several planning and tool-use errors. The system has been productionized and significantly accelerated iteration cycles for the launch of a conversational recommendation agent at Spotify.\footnote{\url{https://newsroom.spotify.com/2026-04-20/deeper-music-features-about-the-song-dj-songdna/}}. Online A/B tests demonstrate its effectiveness, with +14\% user listening, +5\% increase in weekly active users, and a 5\% reduction in skip rate compared to a prior experience that only supports session refinement.

Overall, this work provides a practical framework for accelerating the development of conversational recommendation agents in industry, addressing challenges that are becoming increasingly central to recommender systems and agentic applications as natural-language interfaces reach widespread adoption.
\end{abstract}

\maketitle

\section{Introduction}
\begin{figure}[!t]
    \centering
    \includegraphics[width=\linewidth]{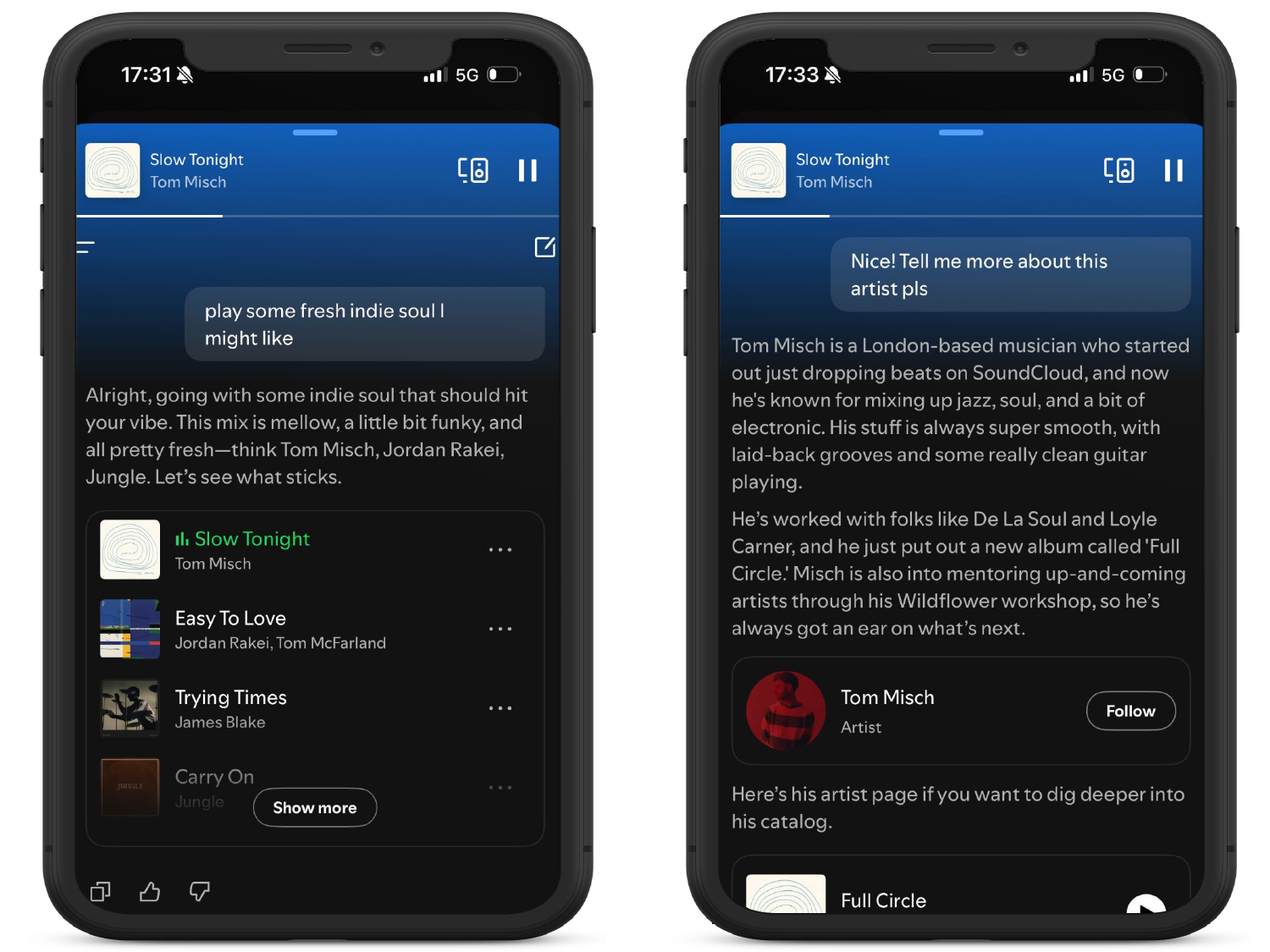}
    \caption{Conversational Recommendation Agents enable users to actively ask for recommendations and continue the dialog using natural language}
    \label{fig:recommendation_agent}
\end{figure}

Recommender Systems are traditionally defined as functions that, given a user profile, retrieve items matching the user’s preferences. This paradigm is often contrasted with search systems, which instead rely on explicit textual queries to retrieve relevant items~\cite{belkin1992information}. This distinction is increasingly blurring with the emergence of \emph{conversational recommender systems}, which jointly model user preferences and natural language input, enabling more expressive and controllable personalization experiences driven by complex, multi-faceted user intents~\cite{jannach2021survey}.

A key advantage of conversational recommenders is their multi-turn nature, which allows users to iteratively refine, adjust, or shift their preferences through follow-up interactions. Moreover, when powered by large language models (LLMs) with tool-use capabilities, conversational recommendation agents can go beyond item recommendation to support richer interactions within the same experience, such as answering questions about content (\emph{``who is this artist?''}) or providing contextual explanations  (\emph{``which artists influenced my favorite artist, and why?''}). 


Realizing this potential, however, requires overcoming key challenges. While traditional recommender systems benefit from abundant user interaction data, large-scale multi-turn conversational data is typically unavailable, especially prior to product launch. At Spotify, recent systems such as our agentic search architecture~\cite{palumbo2025you} and the DJ\footnote{\url{https://newsroom.spotify.com/2025-05-13/dj-voice-requests/}} experience support a range of user requests, but remain largely limited to single-turn interactions and simple session refinement requests.


In this paper, we address the problem of bootstrapping a conversational recommendation agent in a cold-start setting, where both multi-turn data and reliable evaluation signals are scarce. In such settings, the lack of realistic conversational data makes it difficult to systematically evaluate multi-turn capabilities, while the stochastic nature of agent planning makes failures difficult to diagnose and correct issues. We therefore focus on two key challenges:
(1) generating realistic multi-turn user–agent conversations from existing single-turn data, to enable pre-launch evaluation, and
(2) improving agent planning under stochastic and data-scarce conditions to ensure reliable and consistent tool use.

To address these challenges, we first define a taxonomy of multi-turn conversational capabilities, grounded in prior work~\cite{deshpande2025multichallenge, bai2024mt} and product requirements, together with a pipeline for generating synthetic user–agent conversations. We then develop a self-improvement loop based on a coding agent that uses evaluation variance to distinguish inconsistent planning behaviors from capability gaps. The loop combines contrastive optimization, i.e., comparing successful and unsuccessful agent plans to identify reliable planning patterns, with iterative refinement for harder failure cases.

Our approach produces high-quality multi-turn conversations, with human annotation scores above $90\%$ across all quality dimensions. This data enables fine-grained evaluation of conversational capabilities, revealing that performance varies across capabilities and degrades as conversations grow beyond four turns. The self-improvement loop further improves agent performance by +8\% on top of a highly optimized prompt, while uncovering and resolving subtle planning and tool-use issues. Together, these components significantly accelerate iteration on agent planning and enable reliable development in cold-start settings.

These methods have been instrumental in the launch of a conversational recommendation agent at Spotify. Online A/B tests demonstrate significant improvements in user engagement, which led to its subsequent production deployment (Figure~\ref{fig:recommendation_agent}). More broadly, our results suggest that synthetic-data-driven evaluation combined with automated self-improvement provides a practical and generalizable approach for developing conversational recommendation agents in data-scarce settings.




\section{Related Work}
\label{sec:related_work}
We review prior work in conversational recommendation, conversational data generation, and agent self-improvement.

\paragraph{Conversational Recommendation}
Conversational recommendation is a form of conversational information seeking~\cite{conversationalinfoseeking} defined as \emph{``a software system that supports users in achieving recommendation-related goals through a multi-turn dialogue''}~\cite{jannach2021survey}. The vision dates back to early work that framed recommendation as an interactive preference-elicitation process driven by question-answer exchanges~\cite{christakopoulou2016towards}. Beyond eliciting user needs, conversational recommenders are expected to recommend items, explain those recommendations, and answer follow-up questions~\cite{jannach2021survey}. 

Historically, conversational recommender systems have been built as pipelines of specialized components (e.g., query understanding, item retrieval and ranking, dialogue management, and response generation) coordinated through hand-crafted rule-based logic~\cite{jannach2021survey}. With the advent of LLMs and their general-purpose reasoning abilities, \emph{end-to-end} approaches have emerged along two complementary directions. \emph{Generative retrieval} approaches frame recommendation as the direct generation of item identifiers conditioned on conversational context~\cite{tay2022transformer, rajput2023recommender, palumbo2025text2tracks, he2026plum}. \emph{Agentic} approaches instead allow LLMs to plan and call existing recommendation services as tools, leveraging production-grade retrieval and ranking systems while keeping conversational reasoning within the LLM~\cite{palumbo2025you, doh2025talkplay}. Our work falls into the latter category and focuses on bootstrapping such agents in cold-start settings.

\paragraph{Conversational Data Generation}

Evaluating recommendation agents only in the single-turn setting is insufficient. Recent studies show that even frontier LLMs degrade significantly as conversations grow, getting ``lost'' across turns and failing to maintain consistent grounding and instruction following~\cite{laban2025lost}. Furthermore, key capabilities of conversational agents, such as refining a music session or asking follow-up clarifications,  only arise in a multi-turn setting. This motivates the need for realistic multi-turn data that captures tool use, contextual reasoning, and intent shifts. 

For general-purpose agents, recent benchmarks rely heavily on synthetic or semi-synthetic multi-turn data: MT-Bench-101~\cite{bai2024mt} and MultiChallenge~\cite{deshpande2025multichallenge} probe fine-grained conversational capabilities of frontier LLMs, while $\tau$-bench~\cite{yao2024tau} simulates tool-agent-user interactions and evaluates how agents interact with simulated users. In the more specific setting of conversational music and playlist recommendation, several datasets have been proposed~\cite{chaganty2023beyond,leszczynski2023talk, choi2025talkplaydata}, but are limited to recommendation as the sole intent. In contrast, our pipeline seeds multi-turn conversations from existing single-turn intents and models a broader range of conversational behaviors, including refinement, follow-up questions, and intent shifts.

\paragraph{Agent Self-Improvement}
A growing body of work treats agent prompts and configurations as parameters to be optimized rather than hand-tuned. These approaches aim to improve agent behavior by automatically updating prompts or surrounding components based on feedback. For example, methods such as \emph{TextGrad}~\cite{yuksekgonul2024textgrad} propagate natural-language ``gradients'' through LLM calls, \emph{GEPA}~\cite{agrawal2026gepa} evolves prompts through reflective optimization, and \emph{ACE}~\cite{zhang2025ace} frames self-improvement as context engineering, modifying the input context rather than model weights. Beyond prompts, recent work also optimizes the broader agent \emph{harness}, including tool routing, control flow, and decoding parameters. For instance, \emph{Meta-Harness}~\cite{lee2026metaharness} performs end-to-end optimization of agent pipelines, while \emph{optimize\_anything}~\cite{agrawal2026optimizeanything} provides a general framework for optimizing textual components of LLM systems.

Our self-improvement loop is complementary to these approaches. Rather than proposing a new optimization method, we focus on leveraging \emph{data and evaluation signals} in a production setting, where feedback is often noisy due to infrastructure issues and inherent LLM variability. Concretely, we use synthetic multi-turn conversations to surface failure modes, analyze variance across sampled plans to identify their root causes, and use a coding agent to refine prompts and tool definitions when necessary.

\section{Approach}

We present our approach to multi-turn conversation generation and agent self-improvement, which enable evaluation and optimization of conversational recommendation agents in cold-start scenarios.

\subsection{Multi-turn Conversation Generation}
\begin{figure*}[t]
    \centering
    \includegraphics[width=0.9\linewidth]{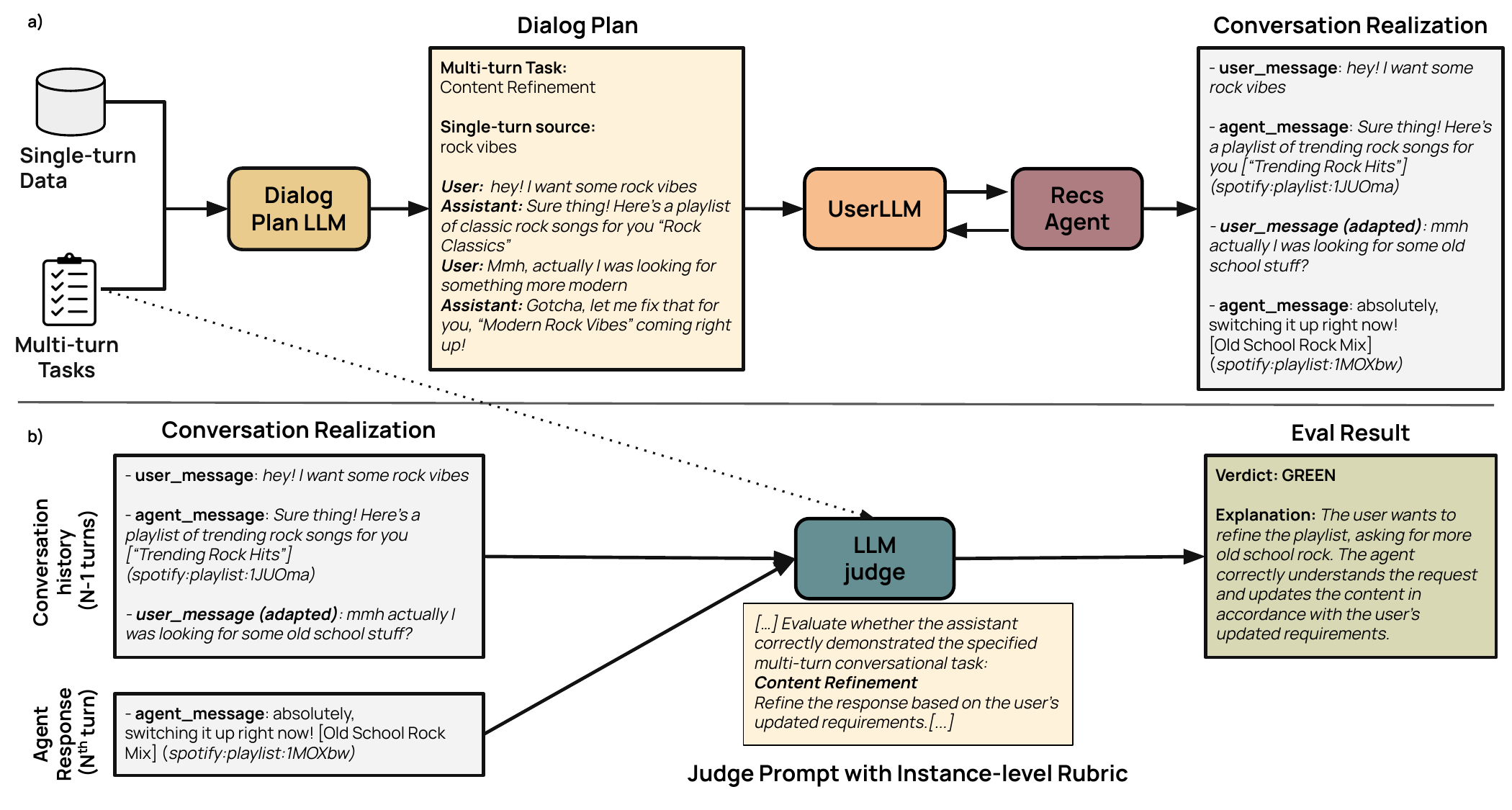}
    \caption{Multi-turn Synthetic Data Generation Pipeline. a) We generalize single-turn prompts into multi-turn synthetic user-agent dialogs. b) The dialogs are then evaluated through LLM-as-a-judge with an instance-level rubric.}
    \label{fig:multi_turn_data_gen}
\end{figure*}

Single-turn evaluation captures whether the agent can resolve a user prompt, but it misses failure modes that emerge as conversations extend across multiple turns. For instance, an agent may retrieve the right type of music for a given request, yet fail to refine it when the user updates their requirements. In a cold-start scenario, real multi-turn traffic with sufficient scale and diversity is not available. We therefore introduce a synthetic generation pipeline that bootstraps
a multi-turn dataset from existing single-turn interactions 
(Figure~\ref{fig:multi_turn_data_gen}).
    
As illustrated in Figure~\ref{fig:multi_turn_data_gen}a, the key design choice behind the pipeline is to separate what the conversation should achieve from how the agent executes it. We first define a 
\emph{dialog plan}, i.e., an abstract sequence of conversational steps designed 
to exercise a target capability, and then \emph{realize} this plan by interacting with the live agent through a UserLLM.

Compared to directly generating user–agent conversations in a single step, this design offers several advantages. First, it provides greater control over the dataset and facilitates versioning, since dialog plans can be audited, stored, and reused across evaluations. Second, it naturally incorporates manually curated multi-turn test cases, which can be represented as dialog plans and adapted to the agent’s responses. Third, it extends naturally to real multi-turn conversations once they become available, which can be logged and reused as dialog plans, with the UserLLM adapting them to the current agent version.


The pipeline consists of five components: taxonomy, single-turn data, dialog plan generation, conversation realization, LLM-as-a-judge. 

\subsubsection{Taxonomy}

Multi-turn conversations can fail in many different ways. To make these failure modes measurable, we define a discrete set of conversational capabilities to evaluate, rather than relying on a vague notion of ``multi-turn quality.'' This allows both data generation and evaluation to target specific conversational capabilities explicitly.


\paragraph{Approach.}
Drawing on prior work~\cite{bai2024mt, deshpande2025multichallenge}, as well as user testing and product priorities, we define a taxonomy
$\mathcal{T} = \{t_1, \ldots, t_5\}$ of five multi-turn capabilities. Each capability 
$t_i = (\textsc{name}_i, \textsc{rubric}_i, \mathcal{E}_i)$ consists of a name, a
natural-language rubric used by the judge for instance-level evaluation, and a small
set of few-shot examples $\mathcal{E}_i$ used to guide dialog plan generation.
The five capabilities are described in Table~\ref{tab:multiturn_taxonomy}.

\begin{table}[t]
\centering
\scriptsize
\setlength{\tabcolsep}{4pt}
\renewcommand{\arraystretch}{1.2}
\arrayrulecolor{black!20}
\begin{tabular}{@{} p{0.36\linewidth} p{0.56\linewidth} @{}}
\toprule
\rowcolor{headbg}
\thead[l]{Task \& Rubric} & \thead[l]{Example dialog plan} \\
\midrule

\rowcolor{rowA}
\textcolor{taskaccent}{\textbf{Content Refinement (CR)}}\newline
\textit{Refine the response based on the user's updated requirements.}
  & \makecell[tl]{%
      \Uturn{} Recommend some relaxing piano playlists. \\
      \Aturn{} Sure---here are a few calming playlists. \\
      \Uturn{} Narrow it down to songs in C major. \\
      \Aturn{} Sure, here are some C major relaxing piano songs.%
    } \\
\midrule

\rowcolor{rowB}
\textcolor{taskaccent}{\textbf{Instruction Retention (IR)}}\newline
\textit{Persist and follow user instructions and preferences across turns (e.g., safety or style constraints) without being reminded.}
  & \makecell[tl]{%
      \Uturn{} Please never include explicit lyrics. \\
      \Aturn{} Understood. I'll only suggest clean tracks. \\
      \Uturn{} Suggest songs for a family picnic. \\
      \Aturn{} Try clean versions of `Levitating' (Dua Lipa).%
    } \\
\midrule

\rowcolor{rowA}
\textcolor{taskaccent}{\textbf{Anaphora Resolution (AR)}}\newline
\textit{Identify pronoun referents throughout a multi-turn dialogue.}
  & \makecell[tl]{%
      \Uturn{} Recommend three artists I should check out. \\
      \Aturn{} Billie Eilish, Justin Bieber, and The Weeknd. \\
      \Uturn{} Who is the third artist? \\
      \Aturn{} The Weeknd is a Canadian singer-songwriter.%
    } \\
\midrule

\rowcolor{rowB}
\textcolor{taskaccent}{\textbf{Intent Shift (IS)}}\newline
\textit{Recognize and focus on the new request when users unpredictably switch intent.}
  & \makecell[tl]{%
      \Uturn{} Recommend some upbeat workout songs. \\
      \Aturn{} Try Calvin Harris or David Guetta. \\
      \Uturn{} Actually, suggest a mindfulness podcast. \\
      \Aturn{} Try ``The Daily Meditation Podcast.''%
    } \\
\midrule

\rowcolor{rowA}
\textcolor{taskaccent}{\textbf{Separate Input (SI)}}\newline
\textit{The first turn outlines the task requirements and the following turns specify the task input.}
  & \makecell[tl]{%
      \Uturn{} Make me a playlist. \\
      \Aturn{} Sure---what kind? \\
      \Uturn{} Bon Iver hits. \\
      \Aturn{} Got it---building a Bon Iver top hits playlist.%
    } \\

\bottomrule
\end{tabular}
\caption{The five multi-turn tasks in our taxonomy. Each task is defined by a natural-language rubric used by the LLM-as-a-judge and illustrated by a representative dialog plan. \Uturn{} and \Aturn{} denote user and assistant turns.}
\label{tab:multiturn_taxonomy}
\end{table}

The taxonomy serves two complementary purposes. During data generation, the examples $\mathcal{E}_i$ guide the
DialogPlanLLM to produce conversations exercising a specific capability $t_i$. During 
evaluation, the rubric $\textsc{rubric}_i$ is incorporated into the judge prompt, ensuring that
scoring is aligned with the intended capability rather than to a generic notion of helpfulness.

\subsubsection{Single-Turn Data}

Single-turn prompts from real traffic are  valuable for capturing the types of requests users are mostly interested in. However, they are inherently limited by the capabilities of existing product surfaces. For example, informational queries are relatively rare because they are not well supported in current products. In this cold-start scenario, we therefore complement real prompts with synthetic ones that reflect product priorities and anticipated conversational behaviors.

\paragraph{Approach.}
The seed corpus $\mathcal{S}$ is defined as the union $\mathcal{S} = \mathcal{S}_{\text{real}}
\cup \mathcal{S}_{\text{synth}}$, where $\mathcal{S}_{\text{real}}$ is sampled from
existing Spotify
product traffic,\footnote{\url{https://newsroom.spotify.com/2025-05-13/dj-voice-requests/}}
and $\mathcal{S}_{\text{synth}}$ consists of synthetic prompts targeting  underrepresented but important use cases. These seed prompts are then used by the DialogPlanLLM to generate multi-turn conversation plans. The resulting training/validation/test split of $\mathcal{S}$ is propagated to the multi-turn dataset to prevent leakage across splits at the seed level.

\subsubsection{Dialog Plan Generation}

A straightforward way to obtain multi-turn dialogs is to roll out a user simulator against the agent and rely on emergent behavior. In practice, this produces conversations that vary across runs, are difficult to reuse, and depend on the specific agent version.

Instead, we first define an abstract
\emph{dialog plan}, i.e.,  a turn-by-turn script designed to exercise a specific conversational capability, and only then realize it through interaction with the live agent. This separation provides greater control over the dataset, enables reuse across evaluations, and ensures that each generated conversation is grounded in a well-defined target capability  $t_i \in \mathcal{T}$. Moreover, dialog plans can be naturally extended with manually curated examples or real user conversations as they become available.

\paragraph{Approach.}
Given a conversational capability $t_i \in \mathcal{T}$, a seed prompt $q \in \mathcal{S}$, and a
target conversation length $n_{\text{turns}}$, the DialogPlanLLM generates a candidate conversation
\[
  d = \big( (u_1, a_1),\, (u_2, a_2),\, \ldots,\, (u_{n_{\text{turns}}},
            a_{n_{\text{turns}}}) \big),
\]
where $u_k$ and $a_k$ denotes the user and agent turns at step $k$. 

The  DialogPlanLLM is conditioned on the task rubric $\textsc{rubric}_i$ and the few-shot examples
$\mathcal{E}_i$, and is instructed to generate a conversation exercising the target capability $t_i$. In particular, the first $n_{\text{turns}} - 1$ provide the conversational context, while the final 
turn $(u_{n_{\text{turns}}}, a_{n_{\text{turns}}})$ defines the behavior under evaluation. 
For example, given the Content Refinement capability and the seed \emph{``rock vibes,''} the DialogPlanLLM may generate a two-turn conversation in which the agent
first returns a playlist and the user then asks for a more modern version  (Figure~\ref{fig:multi_turn_data_gen}a).

To cover a range of conversation lengths, we sample  $n_{\text{turns}} \sim \mathcal{U}\{2, \ldots, n_{\max}\}$
for each generated conversation. The target length $n_{\text{turns}}$ is provided to the DialogPlanLLM as a soft constraint, meaning that it guides generation rather than being enforced through truncation. The train/validation/test assignment of each generated conversation inherits that of its seed prompt $q$ in $\mathcal{S}$, ensuring consistency across splits.

\subsubsection{Conversation Realization}

In a dialog plan, the agent turns $a_k$ are
synthetically generated by the DialogPlanLLM rather than by the live conversational recommendation agent, and therefore lack the
structured outputs (e.g., playlist URIs, album URIs, tool outputs) present in real interactions. Simply replaying the dialog plan would therefore evaluate
the agent in an unrealistic and overly simplified setting. To address this, we preserve the \emph{structure} of the dialog plan , i.e., the sequence of interactions designed to exercise a target conversational capability, while grounding each turn in the agent's actual behavior.

\paragraph{Approach.}
We realize the dialog plan through a closed-loop interaction between the UserLLM and the live agent. The first user turn $u_1$ is sent
verbatim to the agent, which produces a real response
$\hat{a}_1$ containing both text and structured outputs. For each subsequent turn $k > 1$,
the UserLLM generates the next user message $\hat{u}_k$ conditioned on:
\begin{itemize}
    \item the original dialog plan $d$, which specifies the target conversational capability,
    \item the observed conversation history
          $\big( (u_1, \hat{a}_1), \ldots, (\hat{u}_{k-1}, \hat{a}_{k-1}) \big)$,
    \item and the planned user turn $u_k$, which serves as a reference but may be
          adapted.
\end{itemize}
The UserLLM is instructed to follow the plan whenever it remains consistent with 
the agent’s previous response $\hat{a}_{k-1}$, and to minimally adapt $u_k$ otherwise, while preserving the intended conversational capability. 
For example, in a Content Refinement scenario, if the planned user turn is \emph{``make it more modern''} but the agent has already
returned a modern playlist, the UserLLM may instead rewrite the request as 
\emph{``make it more old school''} to preserve a meaningful refinement behavior (Fig.~\ref{fig:multi_turn_data_gen}a). The resulting realized conversation
$\hat{d} = \big( (u_1, \hat{a}_1), \ldots, (\hat{u}_{n_{\text{turns}}},
\hat{a}_{n_{\text{turns}}}) \big)$ preserves the intended capability while remaining grounded
in the agent's actual behavior.

\subsubsection{LLM-as-a-Judge}
\label{sec:judge}

Once a conversation $\hat{d}$ has been generated against the live agent, we need a
scalable way to assess whether the agent successfully performed the intended conversational capability and we use an LLM-as-a-Judge approach~\cite{rahmani2024llmjudge,zheng2023judging}. To have a more fine-grained assessment of multi-turn behavior, we use an instance-level
rubric tied to the specific capability $t_i$ used to generate the dialog. This evaluation process is illustrated in Figure~\ref{fig:multi_turn_data_gen}b.

\paragraph{Approach.}
For a realized conversation $\hat{d}$ generated under capability $t_i$, the LLM-as-a-judge
receives the first $n_{\text{turns}}-1$ turns as conversational context and evaluates only the final agent turn $\hat{a}_{n_{\text{turns}}}$. The
judge's prompt is conditioned on the rubric $\textsc{rubric}_i$, ensuring that evaluation is aligned with the intended capability. 

The judge returns
a label $ y \in \{\textsc{green}, \textsc{red}\}$, where \textsc{green} indicates that the final turn satisfies the rubric and \textsc{red} indicates a failure, along with a free-form explanation. While these explanations are not used in aggregate metrics, they are retained for error analysis and to support the self-improvement loop described  next. 

An alternative approach to evaluating only the final agent turn would be to score every turn in the conversation. However, this would increase evaluation cost and latency linearly with the number of turns. Instead, we focus on the final turn while varying conversation lengths across the dataset, allowing us to assess multi-turn performance at different conversational horizons in aggregate.
\subsection{Self-Improvement Loop}
\label{sec:self-improvement}
\begin{figure*}
    \centering
    \includegraphics[width=0.9\linewidth]{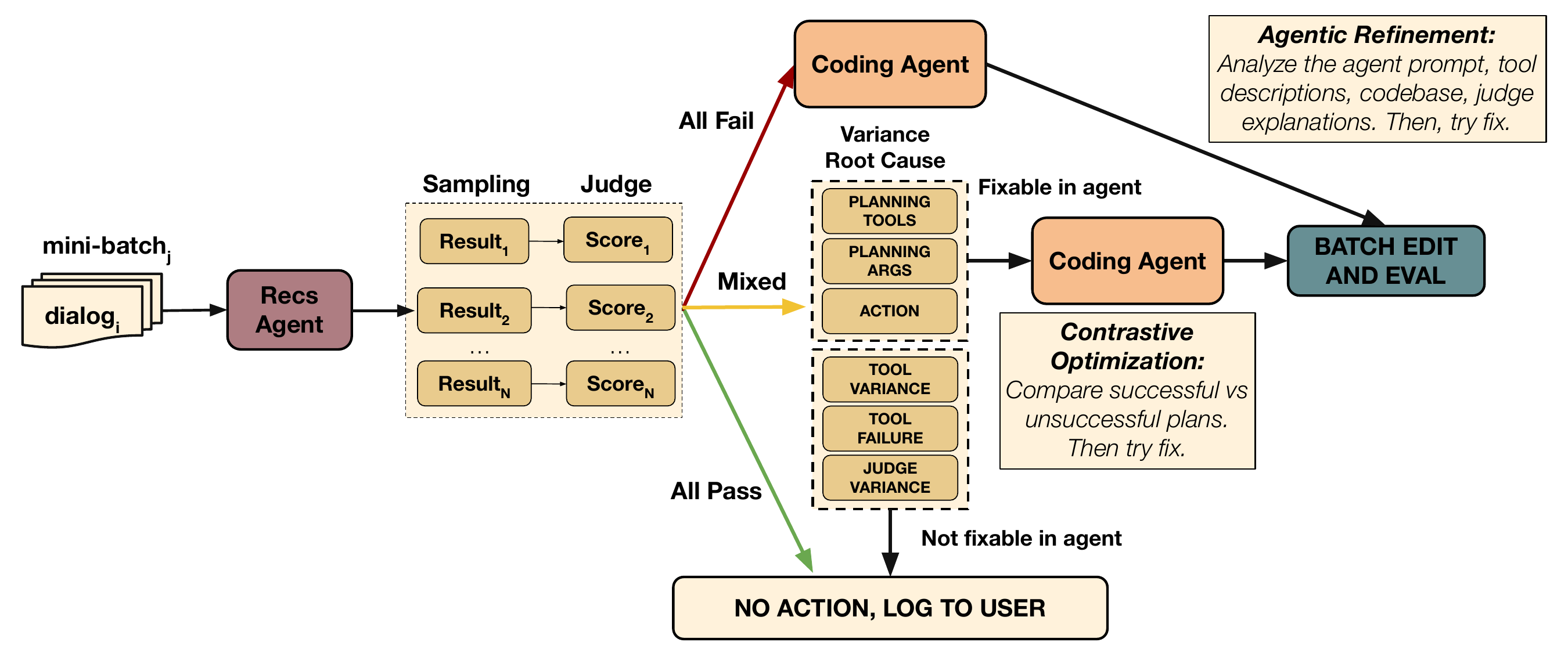}
    \caption{Agent Self-Improvement Loop. We setup a process that leverages the variance in agent planning and the ability of coding agents to diagnose issues and propose improvements to self-improve the agent prompt and tools.}
    \label{fig:placeholder}
\end{figure*}
Once synthetic data has been generated and validated, the remaining challenge is how to efficiently optimize agent planning using this data. In practice, prompt engineering for conversational agents is a manual and iterative process: given a failing query, one inspects the execution trace, identifies the root cause, modifies the prompt or tool descriptions, and re-runs evaluation to verify the fix. This process is time-consuming and does not scale.

We therefore introduce a self-improvement loop inspired by recent work on prompt and agent harness optimization (Section~\ref{sec:related_work}). As illustrated in Figure~\ref{fig:placeholder}, the loop repeatedly samples agent plans, evaluates them using the LLM-as-a-judge, attributes the source of failures through variance analysis, and applies targeted improvements through a coding agent.


A key observation motivating the loop is that not all failures are of the same nature. In practice, some failures correspond to inconsistent planning behaviors that occasionally succeed under sampling, while others reflect genuinely missing capabilities that cannot be recovered through exploration alone.

Formally, let
$\text{pass}@k$ denote the probability that at least one of $k$ independent sampled plans is
correct (i.e., rated as GREEN by the judge), and let $\text{pass}^ k$ denote the probability that \emph{all} $k$ sampled plans are correct. We estimate these quantities using unbiased estimators following~\cite{yao2024tau, chen2021evaluating}. 
Empirically, we observe a large gap between them (e.g., $32\%$ at $k=5$), which reveals  two distinct failure modes:
\begin{itemize}
  \item \textbf{Reliability failures} ($\text{pass}@k \gg \text{pass}^k$, $\text{pass}@k > 0$):
        the correct behavior exists in the model’s output distribution but is not consistently selected. 
  \item \textbf{Capability gaps} ($\text{pass}@k \approx 0$): the correct behavior is absent from the model’s distribution and cannot be recovered through sampling alone.
\end{itemize}

These modes require different treatment. As shown in Figure~\ref{fig:placeholder}, reliability failures are handled through contrastive optimization over successful and unsuccessful plans, while capability gaps trigger iterative refinement through a coding agent. 


However, reliability failures are often noisy and may arise from multiple sources. Variance may stem from genuinely different plans, variations in tool arguments, inconsistencies in tool outputs, or even disagreement from the judge itself. We therefore structure the loop into gove stages: (i) sampling; (ii) root-cause detection; (iii) contrastive optimization for reliability failures; and (iv) agentic refinement for capability gaps (v) batch editing and evaluation.


The self-improvement loop is implemented through a coding agent equipped with read–write access to the prompt artifacts (e.g., system prompt, few-shot examples, tool descriptions), and read access to full execution and evaluation traces (conversation, agent plan, tool calls, tool outputs, and judge feedback). Auxiliary scripts support trace parsing and variance analysis throughout the process.

\subsubsection{Sampling}
The first step in our self-improvement loop is to distinguish reliability failures from capability gaps. To do that, we need to sample multiple responses for the same query and run our evaluation.

\paragraph{Approach.}
As illustrated in Figure~\ref{fig:placeholder}, we begin by sampling multiple agent plans for the same query $q$ at elevated temperature
$\tau \in [0.7, 1.0]$:
\[
  \pi_i \sim p_\theta(\cdot \mid q,\, \mathcal{P}), \quad i = 1, \ldots, k,
\]
where $\mathcal{P}$ denotes the current prompt configuration (system prompt, tool descriptions, and few-shot
examples). Each sampled plan $\pi_i$ is executed and evaluated by the LLM-as-judge (Section~\ref{sec:judge}, producing a 
label $y_i \in \{\textsc{green}, \textsc{red}\}$). This yields two sets of executions:
\[
  \mathcal{D}^+ = \{\pi_i : y_i = \textsc{green}\}, \qquad
  \mathcal{D}^- = \{\pi_k : y_k = \textsc{red}\}.
\]
If all samples belong to $\mathcal{D}^+$, the query is consistently successful and no action is needed (`all pass'). If some samples belong to $\mathcal{D}^+$ and some belong to $\mathcal{D}^-$ (`mixed'), we proceed with the Root Cause Detection step and Contrastive Optimization. If all samples belong to $\mathcal{D}^-$, we go directly to Agentic Iterative Refinement (`all fail').

\subsubsection{Root Cause Detection}

Variance in agent evaluation results can arise from multiple sources. Some reflect meaningful signal for agent self-improvement, while others are simply noise. Treating all sources of variance uniformly would be both inefficient and potentially harmful, increasing the risk of incorrect updates.
A prerequisite for variance-based optimization is therefore to attribute each instance of variance to its root cause.

\paragraph{Approach.}
 
We compare successful and failing executions to identify the source of variance using a combination of trace parsing and LLM-based reasoning. The resulting root-cause taxonomy includes:
\begin{itemize}
  \item \textsc{planning\_tools} — different tool selection
        (e.g.\ \texttt{SearchTool} vs \texttt{RecommendationTool}).
  \item \textsc{planning\_args} — identical tools but different arguments
        (e.g.\ \texttt{"mk gee top hits"} vs \texttt{"mk gee fresh songs"}).
         \item \textsc{action} — different high-level actions (e.g., playback vs. saving to library)
  \item \textsc{tool\_failure} — identical plan but downstream tool failure (e.g., timeouts, connection issues).
  \item \textsc{tool\_variance} — identical plan and execution  but
        non-deterministic tool outputs
  \item \textsc{judge\_variance} — identical execution but inconsistent judge outputs
\end{itemize}

Deterministic traces features (e.g., tool names, argument hashes, exit codes) are
extracted programmatically, while the LLM is used only for cases requiring semantic interpretation.
%

The root-cause label determines how each execution is handled in the self-improvement loop (Figure~\ref{fig:placeholder}). Only failures attributed to 
\textsc{planning\_tools}, \textsc{planning\_args}, or \textsc{action} are passed to the contrastive optimization stage. Other failure modes that are unrelated to agent planning are logged separately for analysis.

\subsubsection{Contrastive Optimization}

When $\text{pass}@k > 0$, the model is already capable of producing the correct plan, but does so inconsistently. The goal of contrastive optimization is therefore to increase the probability of selecting successful planning behaviors, converting stochastic success into more reliable execution. 

\paragraph{Approach.}
Given a query $q$ whose variance has been attributed to agent planning, the coding agent compares successful and unsuccessful plans from $\mathcal{D}^+$ and $\mathcal{D}^-$ to identify the planning patterns associated with success. These patterns typically involve differences in tool selection, tool ordering, or argument construction. The coding agent then proposes a prompt improvement $e^*$ that reinforces the successful planning behavior. The updated prompt is 
$ \mathcal{P}' = \mathcal{P} \cup \{e^*\}$.

\subsubsection{Agentic Iterative Refinement}
When $\text{pass}@k = 0$ for all $k$, the model never produces the correct plan, regardless of sampling. In this mode, the required behavior lies outside the support of the current model distribution, and sampling-based optimization is therefore insufficient. Instead, we rely on a coding agent to analyze failures and propose targeted fixes based on the evaluation traces and prompt artifacts.

\paragraph{Approach.}
 Given a consistently failing execution trace for a query $q$, the coding agent analyzes why the current plan $\hat{\pi}$ fails and what alternative plan $\pi^*$ could succeed. The agent then proposes a minimal modification to the prompt configuration $\mathcal{P}$, targeting the specific source of failure (e.g., adding a tool-use example, extending a tool description, or introducing a routing instruction).

Unlike contrastive optimization, where successful plans can already be surfaced through sampling, iterative refinement requires the coding agent to reason about missing capabilities directly. As a result, each iteration is more computationally expensive and requires more reasoning from the agent. We therefore reserve this stage for queries where contrastive optimization has been attempted and exhausted (i.e., $\text{pass}@k = 0$). 

\subsubsection{Batch editing and validation}
To reduce the risk of overfitting, the agent operates on random
\emph{mini-batches} of queries. Given a batch
$Q = \{q_1, \ldots, q_m\}$, the coding agent produces a single update $\Delta\mathcal{P}$ intended to resolve all the fixes identified in the entire batch:
\[
  \Delta\mathcal{P}^* = \arg\min_{\Delta\mathcal{P}}\;
    \frac{1}{m} \sum_{j=1}^{m} \mathcal{L}(q_j,\, \mathcal{P} + \Delta\mathcal{P}),
\]
where $\mathcal{L}(q, \mathcal{P})$ denotes the ``judge loss'' for query $q$ under prompt configuration $\mathcal{P}$. 

Batching encourages the coding agent to identify shared structural causes rather than overfitting to individual queries. 
After each update, we re-run the evaluation with temperature $\tau=0$ to deterministically verify the effectiveness of the modification. If successful, the update is committed and the process continues with the next batch. At the end of the loop, a pull request is opened for human review and a final full evaluation is performed as a guardrail against potential regressions.\\

\section{Results}
\label{sec:results}

In this section, we present results on the multi-turn synthetic data generation pipeline, the self-improvement loop, and the online performance of the conversational recommendation agent.

The following experimental results are obtained using top-tier frontier models with strong off-the-shelf capabilities for all the tasks described, i.e. synthetic data generation, LLM-as-a-judge, coding agent, and for the conversational recommendation agent itself.

\subsection{Multi-turn Evaluation Results}

We first evaluate the quality of the synthetic multi-turn data generated by our pipeline, and then use it to analyze the agent’s conversational capabilities.

\begin{figure}
    \centering
    \includegraphics[width=\linewidth]{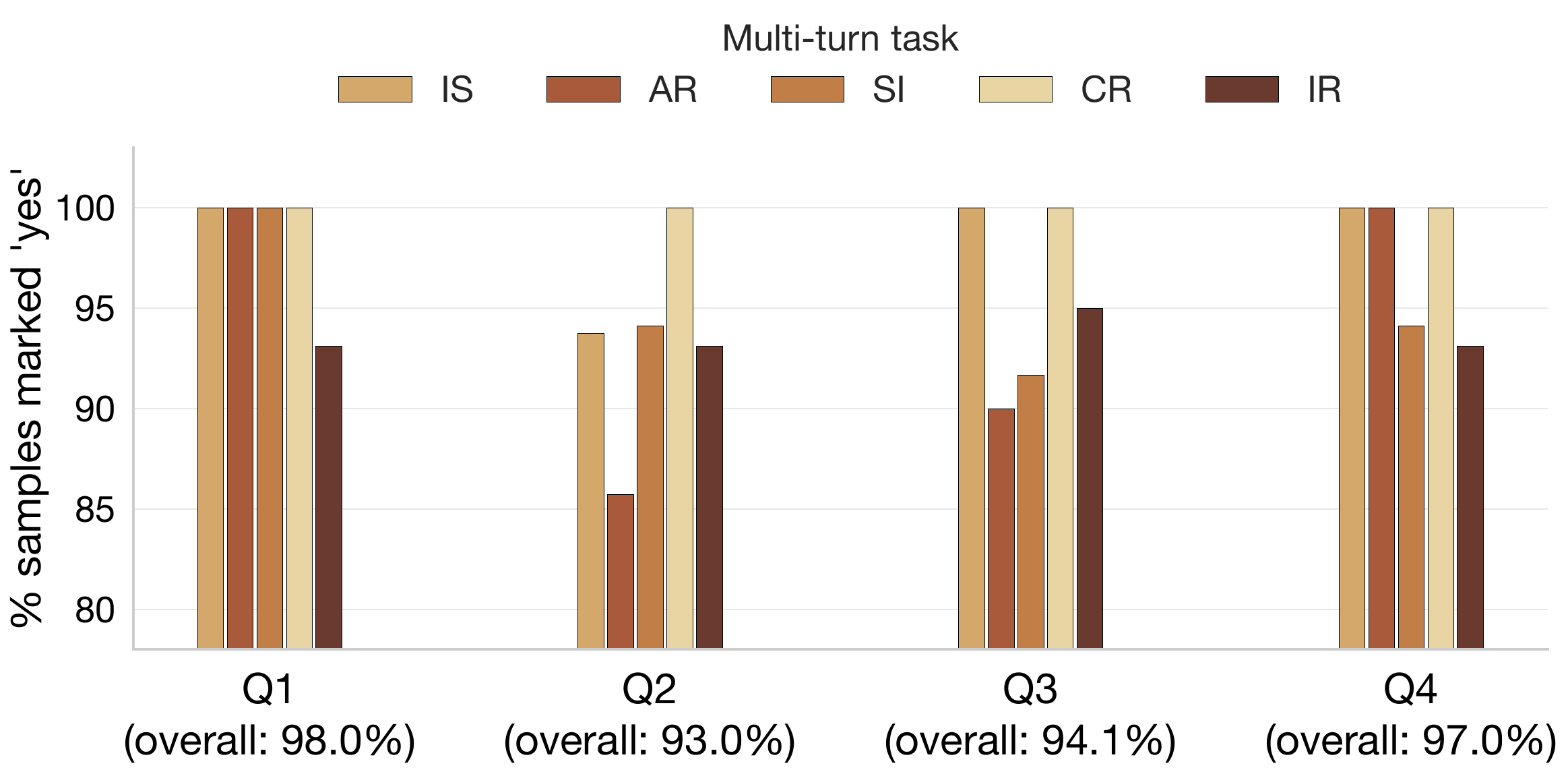}
    \caption{Quality of synthetic multi-turn conversations judged by human annotators}
    \label{fig:manual_annotations}
\end{figure}

\paragraph{\bf What is the quality of the synthetic data?}
Before using the synthetic multi-turn data for evaluation and self-improvement, we benchmark and validate its quality through human annotation. We sample 100 synthetic user-agent conversations stratified by conversational capability and ask five annotators to evaluate each conversation along four binary dimensions:   
  \begin{itemize}    
    \item \textbf{Q1}~(task demonstration): \emph{does the conversation exercise the intended conversational capability, i.e. the multi-turn task?}
    \item \textbf{Q2}~(conversation realism): \emph{is the conversation natural, i.e., could it plausibly come from a real user?}
    \item \textbf{Q3}~(adaptation quality): \emph{when the UserLLM adapts a turn, is the adaptation reasonable given the original dialog plan? }
    \item \textbf{Q4}~(judge accuracy): \emph{is the judge verdict correct and  aligned with human judgment?}
  \end{itemize}                                                                                                       
Results are shown in Figure~\ref{fig:manual_annotations}. All four dimensions score above 90\%, indicating strong overall quality of the synthetic data generated by our pipeline.                                     %
Conversations almost always exercise the target capability (Q1: 98\%). The only capability below 95\% is Instruction Retention, where some conversations repeat the user's constraint across turns, effectively reminding rather than testing retention. 

Conversations are also highly realistic (Q2: 93\%). Most failures occur when the agent's response diverges from the dialog plan, causing the UserLLM to generate an awkward adaptation linking Q2 failures directly to Q3. Overall, UserLLM adaptation quality is strong with 94\% of applicable cases rated as reasonable (Q3). Judge alignment with human annotation is also high, with 97\% of samples rated as accurate (Q4). Notably, annotators never overrule \textsc{RED} verdicts, though a few disagreements arise on \textsc{GREEN} cases. 

To further monitor judge alignment over time, we maintain a curated set of approximately $150$ \emph{golden} examples on which we track near-perfect agreement. 
Note that we use different models for the judge and the agent itself to avoid same model biases.

\paragraph{\bf What are the most challenging conversational capabilities?} 
We analyze the agent’s multi-turn capabilities by  measuring $pass@1$ through LLM-as-a-judge verdicts (i.e., the GREEN rate) across conversational capabilities. Results are averaged over $n=10$ runs to estimate variability.

We find that Intent Shift is the easiest capability, with the agent reliably recognizing changes in user intent during a conversation. Anaphora Resolution performs next best, with the agent correctly identifying references to items mentioned in previous turns (\emph{e.g., ``play the second one''}). Separate Input follows, while Content Refinement and Instruction Retention are the most challenging capabilities. The latter is particularly difficult, as it requires enforcing constraints introduced early in the conversation across turns.

Beyond these aggregate results, multi-turn evaluation also helped identify several failure modes before deployment. 
For instance, in Anaphora Resolution, we observed cases where the agent returned incorrect answers when users referred to playlist items by position. This behavior was traced to incorrect tool-call arguments that altered track ordering.

In Instruction Retention, we identified cases where multiple constraints were partially lost, with the agent \emph{over-summarizing} the prompt during music session creation. Similarly, in Content Refinement, the agent sometimes failed to satisfy constraints due to missing or incomplete metadata. These issues were subsequently addressed through manual fixes and automated improvements using the self-improvement loop described in Section~\ref{sec:self-improvement}.


\begin{figure}
    \centering
    \includegraphics[width=0.7\linewidth]{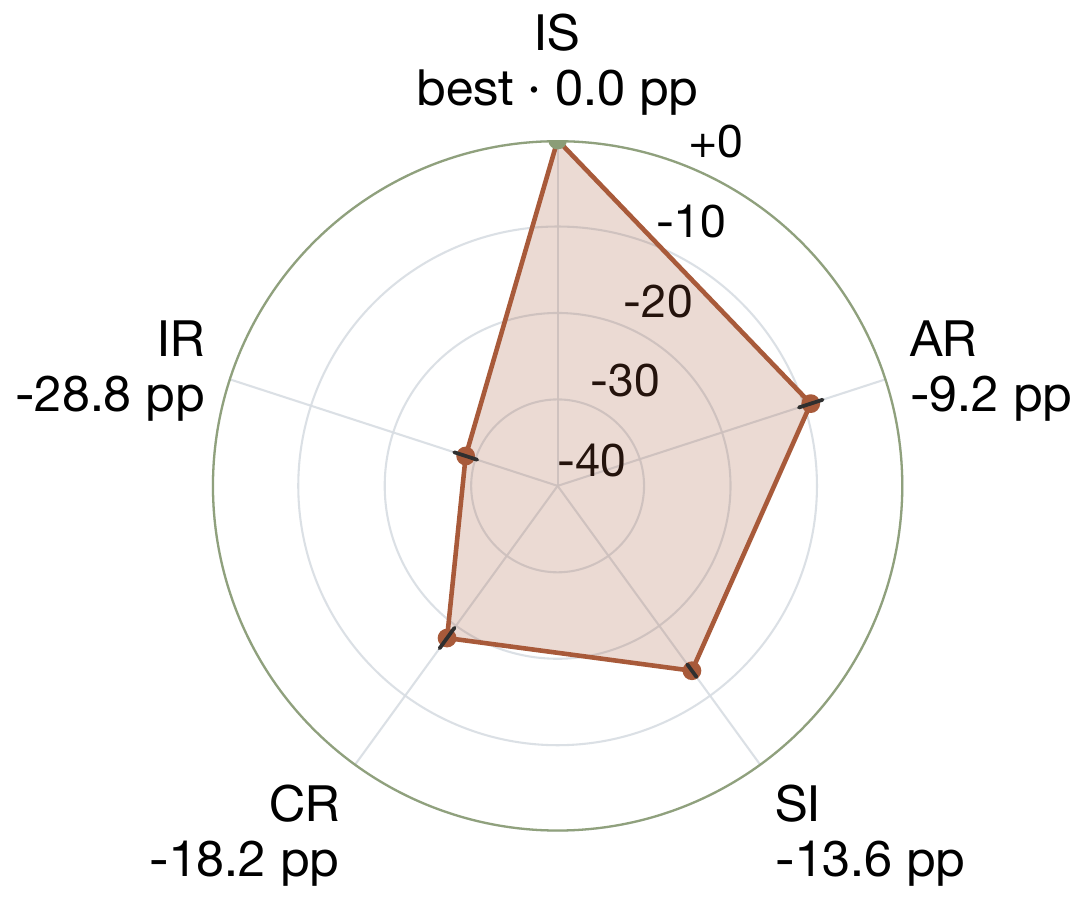}
    \caption{Multi-Turn capabilities vary by task. Relative $pass@1$ with respect to the best task Intent Shift. See definition of each task in Tab.~\ref{tab:multiturn_taxonomy}.}
    \label{fig:mt_tasks}
\end{figure}

\paragraph{\bf How does multi-turn quality vary with conversation length?}
We analyze how $pass@1$ changes as conversation length increases. Performance remains stable from 2 to 4 turns, but degrades from 5 turns onward, consistent with prior findings~\cite{laban2025lost}. Improving long-context handling through techniques such as context compaction and history summarization~\cite{kang2025acon} remains future work.

\begin{figure}
    \centering
    \includegraphics[width=\linewidth]{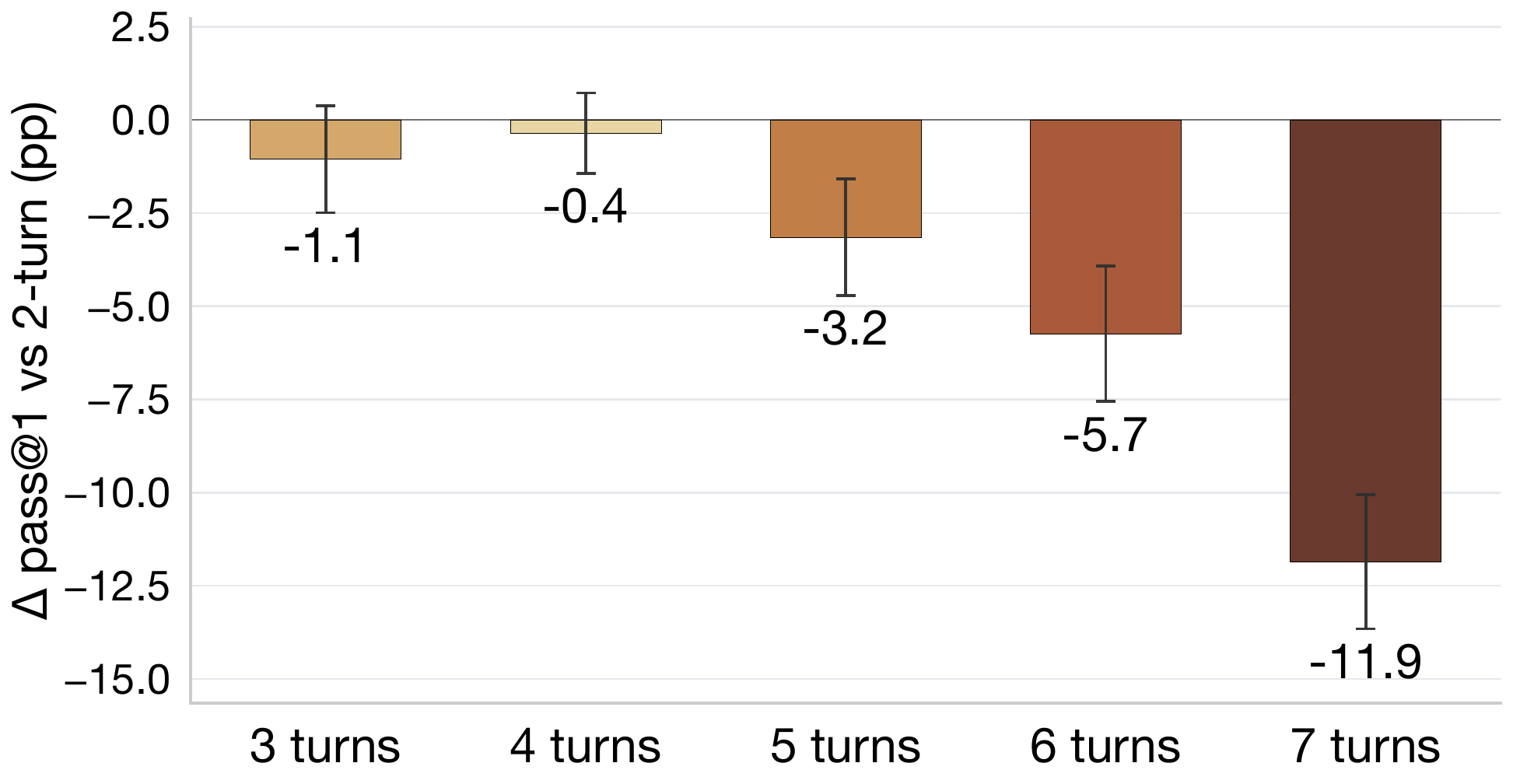}
    \caption{Multi-Turn capabilities vary by conversation length. Relative $pass@1$ with respect to 2-turns conversations.}
    \label{fig:mt_length}
\end{figure}

\subsection{Self-Improvement Results}

We evaluate the effectiveness of the self-improvement loop in automatically improving agent planning at scale.

Given a dataset with approximately $1000$ examples, including both single-turn and multi-turn conversations, we run the self-improvement loop to automatically identify and fix outstanding issues. The data is processed in mini-batches of size 8 to avoid overloading the coding agent with too many issues at once. 
The pipeline typically runs for a few hours, iteratively fixing issues and committing updates to a code branch. 

In one  run, the combined effect of these changes led to +8\% relative increase in $pass@1$ on top of a highly manually optimized prompt on the full test set (statistically significant with $p=0.01$ using a pairwise sign test). 
Examples of identified prompt improvements include:
\begin{itemize}
    \item \textbf{intent clarification in multi-turn intents}: \emph{when a user says `give me more artists' after asking for a playlist, refine the existing playlist adding more variety rather than returning a list of artists}
    \item \textbf{genre disambiguation for music session}: \emph{When a genre name is ambiguous, disambiguate based on conversational context. For instance, if paired with rock genres like `alt rock', `hardcore' means hardcore punk and not electronic hardcore.}
    \item \textbf{time constraints}: \emph{`Year in review', `my year in music' are ALL calendar-year queries - use listening history tools with start time end time for the full calendar year}
    \item \textbf{item type validation}: \emph{if the user is asking for `latest album' but tools returned ONLY singles and no actual albums, you should plan again with the correct entity type filter}
\end{itemize}

All changes proposed by the self-improvement loop are reviewed before being merged into the codebase. In practice, the loop significantly accelerates agent planning optimization. Compared to alternative approaches such as ACE~\cite{zhang2025ace}, we find that contrastive optimization combined with variance-based root-cause analysis helps avoid spurious updates driven by noisy signals (e.g., infrastructure failures, tool variability, or judge inconsistencies).

Finally, using a coding agent rather than a standard LLM for reflection and refinement improves handling of long contexts and complex updates, enabling fixes that go beyond prompt edits, such as improvements to tool descriptions. This is consistent with recent work on harness optimization~\cite{lee2026metaharness}.

\subsection{Online Results}
\label{sec:ab-test}

To validate the effectiveness of the conversational recommendation agent in a production setting, we conducted an A/B test comparing the new experience, which supports chat-like, multi-turn interactions across music session and informational intents, to a baseline limited to session refinement.
The experiment was run over two weeks across several markets, exposing approximately 15 million Spotify users across multiple device types. Online results prove the effectiveness of general purpose conversational recommendation agents: we observed 14\% additional user listening, 5\% increase in weekly active users and a 5\% reduction in skip-rate when interacting with the agent compared to the prior experience.
Based on these results, the conversational recommendation agent was subsequently rolled out in production.

\section{Conclusion and Discussion}

In this paper, we presented an approach to bootstrapping a conversational recommendation agent at Spotify in a cold-start setting. We focused on two key contributions: a synthetic multi-turn conversation generation pipeline and a self-improvement loop that leverages evaluation feedback to iteratively improve agent planning.

We show that the synthetic data pipeline produces high-quality conversations, validated through human annotation with scores above 90\% across all evaluation dimensions. The pipeline enables fine-grained analysis of conversational capabilities, revealing that tasks requiring the manipulation of lists under evolving constraints, such as Instruction Retention and Content Refinement, remain challenging even for frontier models. We also observe that performance degrades as conversations extend beyond four turns.

This result highlights that conversational recommendation agents are subject to the same long-context limitations observed in agents and LLMs more broadly~\cite{laban2025lost}, and motivates techniques for context compaction that preserve key conversational details (e.g., recommended items) while compressing more generic verbal information.

We also describe how the self-improvement loop yields a +8\% quality improvement on top of a highly optimized prompt, while uncovering concrete planning and tool-use issues. Although outputs are always reviewed by humans before deployment, the loop significantly accelerates iteration on agent planning.

More generally, these results align with the emerging trend of using coding agents to automate the improvement of models and agents through iterative ``auto-research'' workflows~\cite{lee2026metaharness, agrawal2026optimizeanything, karpathy2026autoresearch}, demonstrating their feasibility and effectiveness in a real industry setting. In our experiments, contrastive optimization combined with variance-based root-cause analysis proved particularly important for improving the signal-to-noise ratio, reducing hallucinations, and increasing the efficiency of the self-improvement loop.

Finally, online experiments demonstrate the practical impact of our work: the conversational recommendation agent drives more than 14\% additional user listening, increases weekly active users by 5\%, and reduces skip rate by over 5\% compared to the prior session-refinement-only experience.

These results validate both the usefulness of the feature and the effectiveness of our approach, which builds on our previous work on scalable agentic query understanding at Spotify~\cite{palumbo2025you} and extends it to multi-turn conversational recommendation and information seeking. More broadly, this work is part of an ongoing effort to enable natural and conversational exploration of the Spotify catalog through Agentic Search and Recommendations.

Several directions remain open. On the data side, we plan to extend the taxonomy to cover informational and meta-conversational intents and to incorporate real production traffic as it becomes available. On the optimization side, we aim to integrate harness optimization methods~\cite{lee2026metaharness, agrawal2026optimizeanything} and extend self-improvement beyond prompts and tool definitions to the broader agent pipeline.

Overall, we believe the approach presented in this paper provides a scalable and practical framework for evaluating and improving conversational agents in the absence of real user interactions. As natural-language interfaces move from research prototypes to mainstream products, we expect the cold-start challenges addressed here to become increasingly relevant across both recommender systems and agentic applications, and hope this work provides a useful starting point for teams facing similar constraints.


\bibliographystyle{ACM-Reference-Format}
\bibliography{sample_bib}

\end{document}